# Robust Dictionary based Data Representation

REN Wei-Ya

Email: weiyren.phd@gmail.com

(College of Information System and Management, National University of Defense Technology, Hunan Changsha, 410073, China)

**Abstract:** The robustness to noise and outliers is an important issue in linear representation in real applications. We focus on the problem that samples are grossly corrupted, which is also the "sample specific" corruptions problem. A reasonable assumption is that corrupted samples cannot be represented by the dictionary while clean samples can be well represented. This assumption is enforced in this paper by investigating the coefficients of corrupted samples. Concretely, we require the coefficients of corrupted samples be zero. In this way, the representation quality of clean data can be assured without the effect of corrupted data. At last, a robust dictionary based data representation approach and its sparse representation version are proposed, which have directive significance for future applications.

## 1 Introduction

Linear representation method[1], which assumes every sample can be uniquely represented as the linear combination of a sample-dictionary, has recently been a popular method in various applications[2][8][12][13]. Given a set of samples $X = [x_1, x_2, \dots, x_n] \in R^{m \times n}$ and a dictionary $D = [d_1, d_2, \dots, d_k] \in R^{m \times k}$, linear representation aims to find a representation matrix $Z = [z_1, z_2, \dots, z_n] \in R^{k \times n}$ that satisfies

$$X = DZ. \qquad (1)$$

Researches investigate many novel approaches by considering regularizations on linear representation, such as sparse representation (or sparse coding)[3][4][12], nonnegative matrix factorization[5][6][7], graph construction[8][9][15][25][26] and etc.

In linear representation, an important task is to deal with the problem of noise interference. In real applications, the robustness to noise and outliers is a vital evaluating indicator of a approach. Noise is always irregular and hard to estimate in real world. Researches usually assume that noise obeys a certain distribution, such as Gaussian distribution, Uniform distribution and etc. At the meantime, samples are assumed be slightly noise pollution[26] or "sample-specific" corruptions[11][13]. For "sample-specific" corruptions, a reasonable assumption is that corrupted samples cannot be represented by the dictionary. In this paper, we enforce this assumption by leading the coefficients of corrupted samples be zero. In this way, the represented error will contain two parts including corrupted samples and representation error of clean samples. Since only the clean samples can be represented by the dictionary, we can pay more attention to the representation error of clean samples without the influence of corrupted samples. Finally, a robust dictionary based data representation method is proposed to deal with the "sample-specific" corruptions problem. Besides, sparse version of the proposed method is also discussed.

The paper is organized as follows. The proposed robust dictionary based data representation method is presented in Section 2. Then, the sparse version of this method is described in Section 3. Conclusion is given in Section 4.

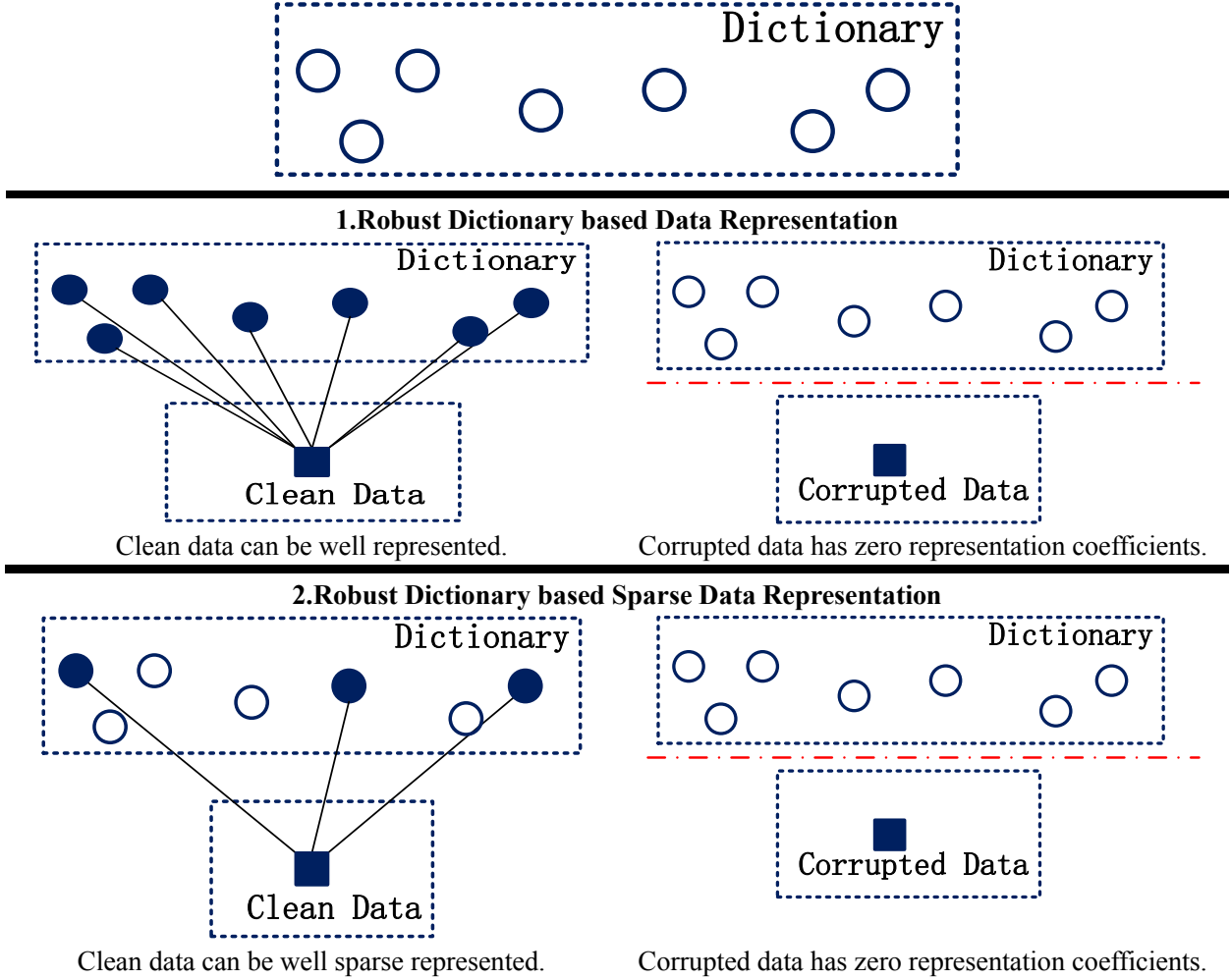

Fig. 1 Dictionary based Data Representation. For robust data representation, clean data can be well represented. At the meantime, corrupted data cannot be represented and its representation coefficients are zero. For robust sparse data representation, clean data can be well sparse represented and corrupted data cannot be represented.

## 2 Robust Dictionary based Data Representation (RDDR)

### 2.1 Dictionary based Data Representation

Given a set $\mathcal{C}$ of samples $X = [x_1, x_2, \ldots, x_n] \in R^{m \times n}$, which $m$ is the dimension of data and $n = |\mathcal{C}|$ is the number of samples. Given an over-complete dictionary matrix $D = [d_1, d_2, \ldots, d_k] \in R^{m \times k}$. Then each column of $Z$ can be represented by a linear combination of $D$. We consider dictionary based data representation as the following optimization problem

$$\min_{Z,E} \quad F_1(Z) + \frac{\lambda}{2} F_2(E)$$
$$s.t. \ X = DZ + E. \tag{2}$$

where $F_1(Z)$ is the regularization term to impose some properties on the representation matrix $Z \in R^{k \times n}$, $F_2(E)$ is the loss function to describe noise $E \in R^{m \times n}$ and $\lambda$ is the tuned parameter.

## 2.2 Definition of $F_1(Z)$

In robust data representation, we can assume that data are noisy and grossly corrupted. To deal with this issue, the state-of-the art data representation methods[13][15][16] define $F_2(E) = ||E||_{2,1}$ to estimate the "sample-specific" corruptions, where $||E||_{2,1} = \sum_{j=1}^{n}\sqrt{\sum_{i=1}^{m}E_{ij}^2}$ is the $l_{2,1}$-norm[11]. $||E||_{2,1}$ encourages the column of $E$ to be zero only when the corresponding column in $X$ is clean.

However, these methods do not have any constraints on the column of $Z$ when the corresponding column in $X$ is corrupted. It is reasonable to assume that a sample cannot be represented by the dictionary if it is corrupted, i.e., the representation of the corrupted sample should be zero. To achieve this goal, we can define

$$F_1(Z) = ||Z||_{q,1} = \sum_{i=1}^{n} ||Z_i||_q. \quad (3)$$

where $Z_i$ is the $i$-th column of $Z$, and $||Z_i||_q = \left(\sum_{j=1}^{m}|Z_{ij}|^q\right)^{\frac{1}{q}}$.

When $q = 2$, $||Z||_{2,1} = \sum_{j}^{n}\left(\sum_{i}^{k} Z_{ij}^2\right)^{1/2}$ if the Group Sparse [18] of the matrix. When $0 < q < 1$, $||Z_i||_q$ is the quasi-norm[19][20] in $\mathbb{R}^K$ space. Specifically, when $q = \frac{1}{2}$, $||Z||_{1/2,1} = \sum_{j}^{n}\left(\sum_{i}^{k} Z_{ij}^{1/2}\right)^2$ is the $l_{1/2}$ sparse of the matrix. When $q \in \{(0,1), 2\}$, $||Z||_{q,1}$ encourages the column of $Z$ to be zero when the corresponding column in $X$ is corrupted and cannot be represented by the dictionary. Generally, $q = 2$ is a common choice.

Then the robust dictionary based data representation problem can be written as

$$\min_{Z,E} \quad ||Z||_{2,1} + \frac{\lambda}{2} F_2(E)$$
$$s.t. \ X = DZ + E. \quad (4)$$

## 2.3 Definition of $F_2(E)$

Now we discuss how to define $F_2(E)$. Suppose the data matrix is $X = [x_1, x_2, ..., x_l, x_{l+1}, ..., x_n]$, where the first $l$ samples are corrupted data and the others are clean samples. Due to the definition of $F_1(Z)$, the ideal $Z$ is $Z = [0, 0, ..., 0, z_{l+1}, ..., z_n]$, and $E$ can be written as $E = [x_1, x_2, ..., x_l, x_{l+1} - Dz_{l+1}, ..., x_n - Dz_n]$. In this way, $E$ can be seen the container of the corrupted samples and the representation error of the clean samples. The first $l$ columns of $E$ are constants and then we only need to care about the estimation of the distribution of the representation error of the clean samples. If we assume the distribution of the representation error of the clean samples is Gaussian distribution, we can define $F_2(E) = ||E||_F^2$. If the clean data still contains some slighted entry-wise corruptions, we can define $F_2(E) = ||E||_1$ ($l_1$-norm). Generally, $F_2(E) = ||E||_F^2$ is a common choice.

Then the robust dictionary based data representation problem can be written as

$$\min_{Z,E} \quad ||Z||_{2,1} + \frac{\lambda}{2} ||E||_F^2$$
$$s.t. \ X = DZ + E. \quad (5)$$

If $Z$ is not ideal, i.e., $Z = [z_1 \to 0, z_2 \to 0, ..., z_l \to 0, z_{l+1}, ..., z_n]$. Then the first $l$ columns of $E$ are larger than other columns of E, and it may affect the distribution estimation of the representation error of the clean samples. One can define $F_2(E) = ||EW||_F^2$ to deal with this problem, where

$W = diag(||Z_1||_q, ..., ||Z_n||_q)$.

Then the robust dictionary based data representation problem can be written as

$$\min_{Z,E} \quad ||Z||_{2,1} + \frac{\lambda}{2}||EW||_F^2$$
$$s.t. \quad X = DZ + E. \tag{6}$$

Another way to deal with the above problem is modeling each column of $E$ as a Mixture of Gaussians (MoG) distribution (MoG distribution is used for modeling noise in [22]).

Same with [22], we assume that each column of $E = [e_1, e_2, ..., e_n]$ follows a Mixture of $s$ Gaussian distributions

$$p(e_i) = \sum_{i=1}^{s} \pi_i \mathcal{N}(e_i|\mathbf{0}, \Sigma_i).$$
$$s.t. \quad \pi_i \geq 0, \sum_{i=1}^{s} \pi_i = 1. \tag{7}$$

where $\pi_i$ denote the mixing weight and $\Sigma_i$ $(i = 1, 2, ..., s)$ denotes the covariance matrix.

Same with [22], we assume all columns in $E$ are independently and identically distributed, then

$$p(E) = \prod_{i=1}^{n} \sum_{i=1}^{s} \pi_i \mathcal{N}(e_i|\mathbf{0}, \Sigma_i). \tag{8}$$

It is equivalent to minimizing

$$F_2(E) = -\ln p(E) = -\sum_{j=1}^{n} \sum_{i=1}^{s} \pi_i \mathcal{N}(e_j|\mathbf{0}, \Sigma_i).$$

Then the robust dictionary based data representation problem can be written as

$$\min_{Z,E} \quad ||Z||_{2,1} - \frac{\lambda}{2} \sum_{j=1}^{n} \sum_{i=1}^{s} \pi_i \mathcal{N}(e_j|\mathbf{0}, \Sigma_i)$$
$$s.t. \quad X = DZ + E, \Sigma_i \in \mathbb{S}^+, \pi_i \geq 0, \sum_{i=1}^{s} \pi_i = 1.$$

where $\mathbb{S}^+$ is the set of symmetrical positive definite (SPD) matrices.

## 2.4 Algorithms

Problems (5) and (6) can be solved by the popular ADM method[13][14]. Problem (5) can also be solved by the linearized alternating direction method with adaptive penalty method (LADMAP)[10]. Notice that problems (5) can be written as

$$\min_{Z} \quad ||Z||_{2,1} + \frac{\lambda}{2}||X - DZ||_F^2. \tag{11}$$

Then we can solve problem (5) by the iteratively reweighted least squares method (IRLS)[17].

Problem (10) can be solved by the EM algorithm[22][23][24]. We list two algorithms in following subsections.

### 2.4.1 Solve problem (5) by LADMAP.

The augmented Lagrange function $\mathcal{L}$ of problem (5) is

$$\mathcal{L} = ||Z||_{2,1} + \frac{\lambda}{2}||E||_F^2 + <Y, X - DZ - E> + \frac{\mu}{2}||X - DZ - E||_F^2. \tag{12}$$

where $Y$ is the Lagrange multiplier, $<\cdot,\cdot>$ is the inner product, and $\mu > 0$ is the penalty parameter.

With some algebra, the updating schemes in each iteration are

$$Z_{t+1} = \underset{Z}{argmin} \ ||Z||_{2,1} + \frac{\mu\eta}{2}||Z - Z_t - D^T(X - DZ_t - E_t + \frac{Y_t}{\mu_t})/\eta||_F^2$$

$$= \Theta_{\frac{1}{\mu\eta}}\left(Z_t + D^T(X - DZ_t - E_t + \frac{Y_t}{\mu_t})/\eta\right). \quad (13)$$

where $\eta = ||D||_2^2$ and $\Theta$ is the $l_{2,1}$ norm minimization operator[13].

$$E_{t+1} = \underset{Z}{argmin} \ \frac{\lambda}{2}||E_t||_F^2 + \frac{\mu}{2}||X - DZ_{t+1} - E_t + \frac{Y_t}{\mu_t}||_F^2 = \frac{\mu_t}{\lambda + \mu_t}(X + \frac{Y_t}{\mu_t} - DZ_{t+1}). \quad (14)$$

The whole algorithm can be seen in algorithm 1.

**Algorithm.1 Solve problem (5) by LADMAP.**

---

**Input**: Dataset $X$, Dictionary data matrix $D$. Parameter $\lambda > 0$.
**Initialize**: $Y = E = Z = \mathbf{0}$, $\mu = 10^{-3}$, $\mu_{max} = 10^{10}$, $\rho = 1.05$, $t = 1$ and $\varepsilon = 10^{-6}$.
**While not converged, do**
    1. Update $Z$ with the other variable fixed
$$Z_{t+1} = \Theta_{\frac{1}{\mu\eta}}\left(Z_t + D^T(X - DZ_t - E_t + \frac{Y_t}{\mu_t})/\eta\right).$$
    2. Update $E$ with the other variable fixed
$$E_{t+1} = \frac{\mu_t}{\lambda + \mu_t}(X + \frac{Y_t}{\mu_t} - DZ_{t+1}).$$
    3. Update the multiplier $Y$ with $Z$ and $E$ fixed
$$Y_{t+1} = Y_t + \mu_t(X - DZ_{t+1} - E_{t+1})$$
    4. Update $\mu$
$$\mu_{t+1} = min(\mu_{max}, \rho\mu_t).$$
    5. Check the convergence conditions
$$||Z_t - Z_{t+1}||_\infty < \varepsilon, \ ||E_k - E_{t+1}||_\infty < \varepsilon, \text{ and } ||X - DZ_{t+1} - E_{t+1}||_\infty < \varepsilon.$$
    6. $t = t + 1$.
**End while**
**Output**: $(Z, E)$.

---

## 2.4.2 Solve problem (5) by IRLS.

By IRLS method, problem (5) can be reformulated as follows

$$\underset{Z}{min} \ \sum_{i=1}^{n}(||Z_i||_2^2 + \mu^2)^2 + \frac{\lambda}{2}||X - DZ||_F^2. \quad (15)$$

where $Z_i$ is the $i$-th column of $Z$.

The derivative of $\mathcal{L}(Z)$ is

$$\frac{\partial \mathcal{L}}{\partial Z} = ZU + \lambda D^T(DZ - X) = 0. \quad (16)$$

where $U$ is a diagonal matrix with the i-th diagonal entry being $U_{ii} = (||Z_i||_2^2 + \mu^2)^{-1/2}$.

Then we can get a well-known Sylvester equation as follows

$$ZU + \lambda D^TDZ = \lambda D^TX. \quad (17)$$

The whole algorithm can be seen in algorithm 2.

**Algorithm.2 Solve problem (5) by IRLS.**

**Input**: Dataset $X$, Dictionary data matrix $D$. Parameter $\lambda > 0$.
**Initialize**: $U = I$, $\mu = 0.1||D||_2$, $\rho = 1.1$, $t = 1$ and $\varepsilon = 10^{-6}$.

1. Update $Z$ by solving the following problem
$$Z_{t+1}U_t + \lambda D^T D Z_{t+1} = \lambda D^T X.$$
2. Update the weight matrix $U$
$$(U_{t+1})_{ii} = (||(Z_{t+1})_i||_2^2 + \mu_t^2)^{-1/2}.$$
3. Update $\mu$
$$\mu_{t+1} = \mu_t/\rho.$$
4. Check the convergence conditions
$$||Z_t - Z_{t+1}||_\infty < \varepsilon.$$
5. $t = t + 1$.

**End while**
**Output**: $Z$.

# 3 Robust Dictionary based Sparse Data Representation (RDSDR)

Since the dictionary matrix are usually over-complete, sparse representation can be considered based on Section two. For clean data, we assume they can be well sparse represented by the dictionary. Meanwhile, we assume corrupted data cannot be represented by the dictionary.

Two ways can be considered based on problem (4), and the objective function can be written as

$$\min_{Z,B,E} \quad ||B||_{1,2} + \beta ||Z||_{2,1} + \frac{\lambda}{2}F_2(E)$$
$$s.t. \quad X = DBZ + E. \tag{18}$$

and

$$\min_{Z,E} \quad ||Z||_1 + \beta ||Z||_{2,1} + \frac{\lambda}{2}F_2(E)$$
$$s.t. \quad X = DZ + E. \tag{19}$$

where $B \in R^{k \times k}$, $||B||_{1,2} = \sum_{i=1}^k \sqrt{\sum_{j=1}^k B_{ij}^2} = ||B^T||_{2,1}$ is the $l_{1,2}$-norm and $||Z||_1 = \sum_{j=1}^n \sum_{i=1}^k |Z_{ij}|$ is the $l_1$-norm[12].

Problem (19) is easy to understand since we hope the representations for clean data are sparse. In problem (18), $B$ can be define as a diagonal matrix and $DB$ become a bases selection process.

Problems (18) and (19) can be solved by the popular ADM method[13][14] and the EM algorithm[22][23][24].

# 4 Conclusion

In this paper, we propose a robust dictionary based data representation method through assuming clean data can be represented by the dictionary while corrupted data cannot. The representation coefficients of corrupted data are designed to be zero. In this way, the representation quality of clean data can be assured without the effect of corrupted data. We extend robust dictionary based data representation (RDDR) to its sparse version (RDSDR). Since there are many ways mentioned before to define of $F_1(Z)$ and $F_2(E)$ respectively, the combination of them may produce various robust dictionary based data representation methods. Furthermore, we can require the representation coefficients are nonnegative for parts-based representation when the dictionary data are nonnegative, i.e., $Z \geq 0$ when $D \geq 0$.